# Pedestrian Behavior Interacting with Autonomous Vehicles during Unmarked Midblock Multilane Crossings: Role of Infrastructure Design, AV Operations and Signaling

**Fengjiao Zou, Ph.D. Candidate (Corresponding author)**
Graduate Research Assistant
Glenn Department of Civil Engineering, Clemson University
S Palmetto Blvd, Clemson, SC 29631
Email: fengjiz@clemson.edu

**Jennifer Ogle, Ph.D.**
Professor and Department Chair
Glenn Department of Civil Engineering, Clemson University
Email: ogle@clemson.edu

**Weimin Jin, Ph.D.**
Traffic and ITS Engineer
Arcadis U.S., Inc.
Email: Weimin.Jin@arcadis.com

**Patrick Gerard, Ph.D.**
Professor
Department of Mathematical Sciences, Clemson University
Email: pgerard@clemson.edu

**Daniel Petty**
Chief Executive Officer
6D systems
Email: daniel@6dsimulations.com

**Andrew Robb, Ph.D.**
Assistant Professor
School of Computing, Human-Centered Computing, Clemson University
Email: arobb@g.clemson.edu

**ABSTRACT**

One of the main challenges autonomous vehicles (AVs) will face is interacting with pedestrians, especially at unmarked midblock locations where the right-of-way is unspecified. This study investigates pedestrians' crossing behavior given different roadway centerline features (i.e., undivided, two-way left-turn lane (TWLTL), and median) and various AV operational schemes portrayed to pedestrians through on-vehicle signals (i.e., no signal, yellow negotiating indication, and yellow/blue negotiating/no-yield indications). This study employs virtual reality (VR) to simulate an urban unmarked midblock environment where pedestrians interact with AVs as they cross a four-lane arterial roadway. Results demonstrate that both roadway centerline design features and AV operations and signaling significantly impact pedestrians' unmarked midblock crossing behavior, including the waiting time at the curb, waiting time in the middle of the road, and the total crossing time. But only the roadway centerline design features significantly impact the walking time. Participants in the undivided scene spent a longer time waiting at the curb and walking on the road than in the median and TWLTL scenes, but they spent a shorter time waiting in the middle of the road. Compared to the AV without a signal, the design of yellow signal significantly reduced pedestrian waiting time at the curb and in the middle. But yellow/blue significantly increased the pedestrian waiting time. Interaction effects between roadway centerline design features and AV operations and signaling are significant only for waiting time in the middle of the road. For middle waiting time, yellow/blue signals had the most impact on the median roadway type and the least on the undivided road. Other factors, such as demographics, past behaviors, and walking exposure of pedestrians, are also explored. Results indicate that older individuals tend to wait longer before making crossing decisions, and pedestrians' past crossing behaviors and past walking exposures do not significantly impact pedestrian walking behavior interacting with AV.

## INTRODUCTION

Between 2011 and 2020, the US witnessed a 46% increase in pedestrian fatalities in motor vehicle crashes, resulting in over 55,000 pedestrian deaths (NHTSA, 2020). In 2020 alone, 6,516 pedestrians were killed in traffic crashes, while approximately 54,769 were injured (NHTSA, 2022). On average, one pedestrian was killed every 81 minutes and injured every 10 minutes in traffic crashes, and pedestrian deaths accounted for 17 percent of all traffic fatalities in 2020 (NHTSA, 2022). Most of these pedestrian fatal and injury crashes occurred in urban areas (82%) rather than rural areas (18%), with 75% of them at midblock locations (NHTSA, 2022). Some researchers found that crossing-related fatal crashes occurred more often in urban/suburban areas on multilane roads at unmarked midblock locations and in roadway sections without median refuge islands (Ogle et al., 2020). Others also found that roads with three or more lanes had a higher pedestrian crash rate than two-lane roads (Zegeer et al., 2001, 2005). These multilane urban/suburban roadways are primarily intended for higher volume and higher speed vehicular mobility, and pedestrians crossing at these locations violate driver expectations.

Recent technological advances like sensor fusion and artificial intelligence have brought the prospect of autonomous vehicles (AVs) closer to reality. AVs have the potential to eliminate crashes caused by human errors and benefit the transportation system (Winkle, 2016). Nevertheless, AVs will face significant challenges, one of which is the gap in social interaction caused by replacing human drivers. For example, interaction with pedestrians presents a challenge. According to some researchers, they are concerned that AVs will be programmed to yield to pedestrians in any situation, even when they cross midblock locations; consequently, pedestrians may perceive that they can cross the road at any time and anywhere, which may force the risk-averse AVs to slow down or stop (Millard-Ball, 2018). Such a risk-averse operating scheme may impede the primary benefit of AVs, namely, transportation system efficiency, while negatively impacting their acceptance. Therefore, understanding pedestrians' behavioral responses under various AV operational scenarios can assist the AVs in responding accordingly.

This study uses virtual reality (VR) to simulate an urban/suburban midblock environment where pedestrians interact with AVs as they cross a four-lane arterial roadway. This study is designed to understand pedestrians' crossing behavior at an unmarked midblock location given different roadway centerline features (i.e., undivided, two-way left-turn lane (TWLTL), and pedestrian refuge island (median)) and various AV operational schemes portrayed to pedestrians through on-vehicle signals (i.e., no signal, yellow negotiation indication, and yellow/blue negotiation/no-yield indications). Outcomes of interest include whether pedestrian behavior changes with the provision of a centerline refuge area or when pedestrians experience different AV operations and signaling. Researchers are also interested in determining interactions between the roadway infrastructure and AV operations and signaling. Other variables, such as demographics, past behaviors, and walking exposure of pedestrians, are also explored.

### Contributions

This paper contributes to previous studies with several unique aspects, including 1) it incorporates pedestrians interacting with AVs on three different multilane roads within a VR setting; 2) pedestrians experience multiple AV interactions in the same scenario as they cross a multilane roadway; 3) AVs are designed to allow real-time, back-and-forth interaction with pedestrians; 4) AVs are designed with some having negotiation behavior, and some having non-yield behavior, and 5) it explores the interaction effects between the roadway infrastructure and the AV operations portrayed through signals.

**LITERATURE REVIEW**

The investigation of pedestrian crossing behavior has been the subject of academic inquiry since the 1950s, with a primary focus on the interaction between pedestrians and human-driven vehicles. Although the analysis of such interactions remains relevant, an increasingly pressing concern pertains to the examination of pedestrian behavior in the context of AVs. Nonetheless, a significant danger arises from conducting empirical testing of pedestrian response to AVs during the development phase. As a result, options for assessing pedestrian response to diverse AV operational schemes are limited. The development of VR technology has enhanced its flexibility and affordability, leading to an increasing number of studies examining pedestrian behavior using VR, particularly in relation to pedestrian interactions with AVs (Schneider & Bengler, 2020). VR-based experiments circumvent the constraints associated with testing AVs in real-world environments, which often necessitate the construction of costly AV prototypes and may pose safety risks to participants. Further, VR has emerged as a promising tool for investigating pedestrian behavior, with research indicating that pedestrian behavior in VR aligns with published real-world norms (Deb, Carruth, et al., 2017). In the context of AV-pedestrian interaction, a significant body of literature has examined the factors influencing pedestrian crossing behavior, including those related to roadway infrastructure, traffic, and vehicle factors.

**Roadway Infrastructure and Traffic Factors**

The majority of studies investigating AV-pedestrian interactions have focused on urban road environments, given the greater prevalence of pedestrian crossing activities in these areas and concerns regarding related crashes (Zegeer & Bushell, 2012). While one study incorporated both rural and urban roads in its design, the impact of these different environments on pedestrian behavior was inconclusive due to the limited sample size of ten participants (Mahadevan et al., 2019). Although multilane roads are common in cities, most studies have employed experimental designs featuring two-lane roads (Ackermans et al., 2020; Deb et al., 2018; Holländer et al., 2019; Jayaraman et al., 2019), with some utilizing one-lane configurations (Camara et al., 2021; Stadler et al., 2019). Additionally, a significant number of studies failed to specify the lane configuration (Tran et al., 2021). One of the primary reasons for employing one or two lanes of roads in VR studies is the physical limitations of the available space (Schneider & Bengler, 2020).

A few studies have examined AV-pedestrian interaction at intersections with crosswalks (Deb et al., 2018; Pillai, 2017; Zhanguzhinova et al., 2023). However, the majority of studies have focused on midblock crossings, with some conducted on marked roads where pedestrians have the right of way (Jayaraman et al., 2018, 2019) and others on unmarked roads where pedestrians do not (Colley et al., 2022; Holländer et al., 2019; Nuñez Velasco et al., 2019). Jayaraman et al. investigated pedestrian crossing behavior at marked midblock crossings and concluded that AVs' aggressive driving behavior had a greater impact on pedestrian behavior at unsignalized crosswalks than at signalized crosswalks, indicating that signalized crossings moderated the negative effects of aggressive driving behavior (Jayaraman et al., 2019). Additionally, Nuñez Velasco et al. employed recorded 360° video in VR to examine pedestrian crossing intentions and found that the presence of a zebra crossing increased pedestrian crossing intention (Nuñez Velasco et al., 2019).

Traffic factors like the number of vehicles, vehicle gaps, and traffic directions also affect pedestrian behavior, and these factors are crucial to the experimental design. Tran et al. reviewed 31 representative studies that investigated AV and pedestrian interaction in VR and found that most studies (94%) only involved one pedestrian interacting with one AV, which limited participants to observing only one vehicle at a time. Although some studies (48%) included multiple vehicles, the interaction remained one-to-one, where the second vehicle appeared only after the first one had been out of the pedestrian's sight (Tran et al., 2021). Only a few studies required participants to interact with multiple vehicles, such as AVs designed to approach from both directions (Colley et al., 2020; Mahadevan et al., 2019). Furthermore, most studies (87%) involved traffic moving on one-way streets, with only a small number (10%) on two-way streets (Tran et al., 2021). For vehicle gaps, research suggests designing multiple experimental conditions with different vehicle gaps (Dietrich et al., 2019). Research showed that a larger gap increased pedestrians' crossing intention (Nuñez Velasco et al., 2019).

**Vehicle Factors**

Vehicle factors have been widely studied and are critical in pedestrian behavior studies within the context of AV driving, as the design of AVs in VR directly influences pedestrian interactions.

Numerous studies have investigated communication between pedestrians and AVs, with a focus on the impact of the external human-machine interface (eHMI) (Rasouli & Tsotsos, 2020; Tran et al., 2021). Böckle et al. conducted a VR-based pedestrian and AV interaction study, concluding that pedestrians' perceived safety and comfort levels were higher when interacting with AVs featuring eHMI compared to those without (Böckle et al., 2017). Other studies reported similar results (Chang et al., 2017; Clercq et al., 2019; Mahadevan et al., 2019). Some researchers also found that eHMI could increase pedestrians' trust (Colley et al., 2020, 2022) and crossing intention (Ackermans et al., 2020; Kooijman et al., 2019). Researchers have also investigated the effects of different types of eHMI. Deb et al. designed four visual and audible AV communication features, with results indicating that these features increased pedestrians' receptivity to AVs (Deb et al., 2018). More specifically, the most favorable visual features were a walking silhouette and a text of "braking", while the most favorable audible feature was a verbal message. Further studies by Deb et al. concluded that children relied entirely on eHMI for crossing decisions (Deb, Carruth, Fuad, et al., 2020), and older pedestrians found eHMI features more important than younger pedestrians (Deb, Carruth, & Hudson, 2020). Similarly, some researchers concluded that a textual display is the clearest (Clercq et al., 2019), while others proved that participants preferred dynamic eHMI over static ones (Othersen et al., 2018). Stadler et al. have demonstrated that eHMI significantly increases pedestrians' reaction times (Stadler et al., 2019). However, potential biases have been noted due to the experimental design that conducts the control group (i.e., without eHMI) first, followed by the randomized five eHMI concepts. But overall, eHMI has been shown to facilitate faster decision-making and earlier initiation of crossing (Ackermans et al., 2020; Chang et al., 2017; Holländer et al., 2019). Some researchers considered that mixed traffic (with different autonomy levels) would become a challenge in the near future and conducted pedestrian interaction with vehicles in mixed traffic conditions in VR, with findings indicating that eHMI is critical in such situations and can increase pedestrians' confidence levels while helping them make crossing decisions faster (Mahadevan et al., 2019). Additionally, researchers have suggested that placing the communication interface on the vehicles is preferable to place them on road infrastructure (Mahadevan et al., 2019).

Some researchers also studied the design of AV driving behavior. But most studies designed AV to drive at a certain speed or yield to pedestrians according to the predetermined deceleration curve. For instance, in the study conducted by Deb et al., AV stopped at the crosswalk for pedestrians in every trial (Deb et al., 2018). The authors also mentioned that this always-conservative AV behavior might encourage pedestrians to cross the road immediately instead of being cautious. Similarly, Chang et al. designed AV to start looking for pedestrians 20 meters away from the crosswalk, start slowdown 15 meters away and stop before the crosswalk (Chang et al., 2017). When no zebra crossing is present, some studies designed AV to never yield to pedestrians (Stadler et al., 2019). Very few researchers have investigated pedestrians' behaviors when interacting with different AV behaviors. Pillai tested pedestrians' comfort levels toward AVs with different deceleration rates and distances to pedestrians and found that early deceleration could provide pedestrians with a higher level of comfort (Pillai, 2017). Other studies also reported similar results (Dietrich, Tondera, et al., 2020; Fuest et al., 2020). Additionally, AVs yielding early to pedestrians can increase traffic efficiency by allowing AVs to accelerate again without coming to a full stop (Dietrich, Maruhn, et al., 2020). Jayaraman et al. designed AV with three different behaviors (defensive, normal, and aggressive) determined by the reaction distance to the pedestrian and the maximum acceleration rate. Results show that aggressive AV behavior that decelerated late decreased pedestrians' trust in AVs (Jayaraman et al., 2018, 2019). Despite these findings, most AV behaviors designed in these studies do not allow for real-time, back-and-forth interaction with pedestrians. Only one study (Camara et al., 2021) continuously adapted AV behavior to pedestrians' movement, which is a more realistic future AV behavior (Tran et al., 2021).

### Research Gap

To summarize the literature above, pedestrian crossing behavior in the context of AV is understudied. It is found that 1) most studies have focused on two-lane roads, with no prior research conducted on multilane roads; also, limited research considered different roadway infrastructure types in their study; 2) the majority of studies have involved traffic moving on a one-way street, and most of the interaction between AVs and pedestrians remained one-to-one in previous studies; 3) most studies have designed simplistic AV behavior without enabling real-time, back-and-forth interaction with pedestrians; 4) very limited research has explored how pedestrians behave differently when interacting with different AV behaviors. Even less attention has been given to the interaction between roadway infrastructures and AV behavior. Therefore, further studies are required to investigate pedestrian crossing behavior given different roadway infrastructure types and AV operational schemes to inform the development of effective AV-pedestrian interaction strategies in diverse urban environments.

### METHODOLOGY

The methodology includes the VR experiment design, data collection, data preparation for modeling, and the statistical model.

### VR Experiment Design

Experiment design in VR includes the design of different roadway infrastructure scenes, different AV behaviors and signals, AV gap design, pedestrian task design in VR, and the data collection procedure. Unity 3D software was used for VR environment development, and Oculus Quest 2, a completely wireless all-in-one VR headset, was used for pedestrian data collection.

*Scenes*

The literature suggests that multilane roads have a higher crash rate compared to two-lane roads (Zegeer et al., 2001). Studies have also demonstrated that the installation of raised medians in multilane locations can result in a safety advantage (Parsonson et al., 2000; Zegeer et al., 2001). In this study, three representative infrastructure scenes of multilane roadways were designed, namely, an undivided 4-lane road, a 4-lane road with a TWLTL in the center, and a 4-lane road with a raised median serving as a pedestrian refuge. Hereafter, these roadway infrastructure scenes will be referred to as the "undivided scene," "TWLTL scene," and "median scene," respectively, as depicted in Figure 1. All roads were designed to be 11 ft each wide, each with a 4 ft sidewalk on both sides. Buildings, walls, and fences were incorporated into the VR design to prevent pedestrians from hitting the physical wall.

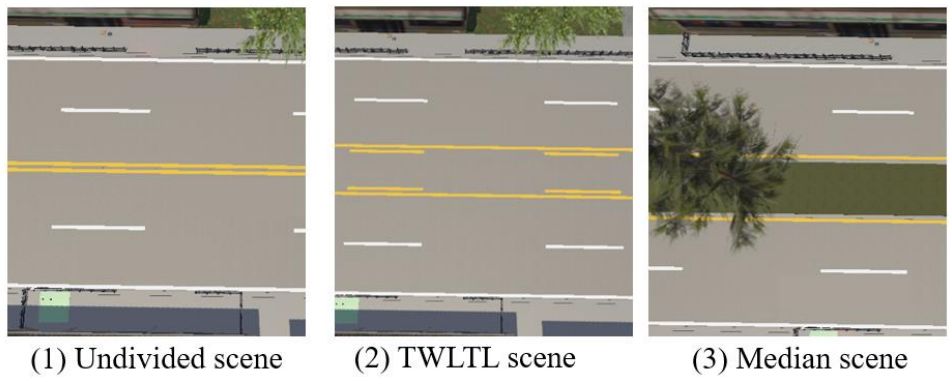

    (1) Undivided scene      (2) TWLTL scene      (3) Median scene

**Figure 1 Roadway Scenes Designed in the VR Experiment**

*AV Behavior and Signals Design*

At unmarked midblock locations, pedestrians do not have the right of way. Several studies have suggested that conservative AV behavior that always stops for pedestrians may negatively impact traffic flow and encourage jaywalking; thus, future AVs may need to engage in negotiations with pedestrians to determine the right of way (Camara et al., 2021; Fox et al., 2018; Gupta et al., 2019; Tran et al., 2021). This study presents a design for AVs that do not always stop for pedestrians at unmarked midblock locations, but instead engage in real-time, back-and-forth interactions with them. AVs were designed to negotiate right-of-way with pedestrians-some may yield, and others will not, based on real-time pedestrians' speed and position. In this study, three AV behavior and signal scenarios included in the experiments are:

- No signal scenario (negotiation behavior with no signal)
- Y signal scenario (negotiation behavior with yellow signal)
- YB signal scenario (a platoon of non-stopping AVs showing blue signals added to the Y signal scenario)

1. No Signal Scenario

This study established a control group without any signals. The AV behavior in this group is identical to that of the Y signal condition depicted in Figure 2, with the sole exception that all signals are deactivated. In the absence of signals, pedestrians are not made aware of whether the AVs have detected their presence or not, and do not receive any communication messages from the AV.

## 2. Y Signal Scenario

Figure 2 presents the Y signal scenario, in which AV negotiates the right of way with pedestrians with a yellow signal. Note that the SSD in Figure 2 represents the stopping sight distance (AASHTO, 2018).

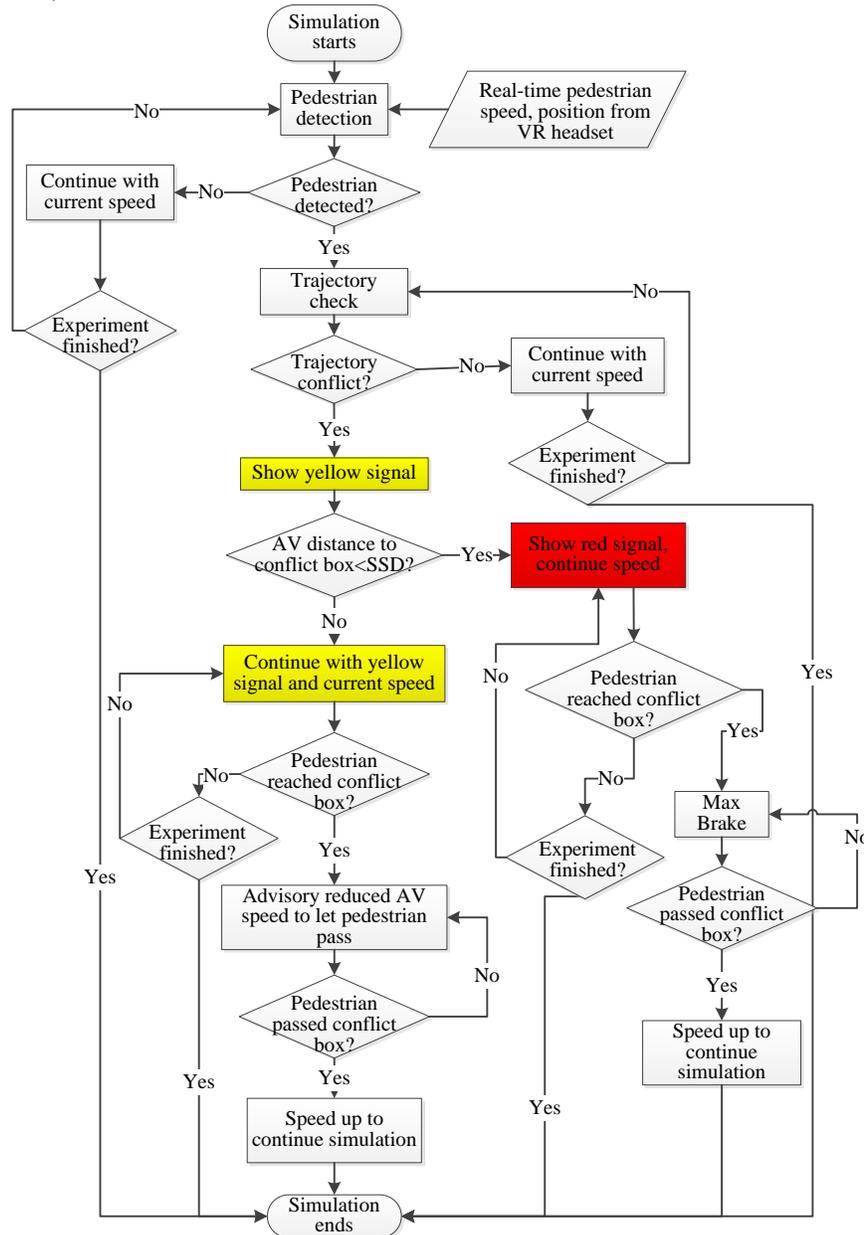

**Figure 2 AV Algorithm with Y Signal Designed**

Figure 2 presents the negotiation process of AV with the Y signal. A trajectory conflict arises between an AV and a pedestrian when the pedestrian is detected at the curb and facing perpendicularly to the AV's direction of travel. Then the AV displays a yellow signal to indicate its intention to negotiate the right of way with the pedestrians. In case the pedestrians remain stationary, the AV continues to move at its regular speed, thereby winning the negotiation. Conversely, if the pedestrians begin to cross, the AV employs real-time monitoring of their position and speed. To facilitate this monitoring, a conflict box, demarcated by a green dashed line in Figure 3 and equal in width to the AV, is established along the pedestrian's path. Upon the pedestrian reaching the conflict box, the AV decelerates and yields to the pedestrians. However, if the pedestrians have not yet reached the conflict box or are hesitating near the curb or edge of the lanes, AVs proceed without slowing down. Notably, the AV emits a red signal as a safety warning when the SSD is less than the AV's distance to the conflict box. The AV speed limit is designed to be 20 mph. This study also did not add turning maneuvers to the AVs.

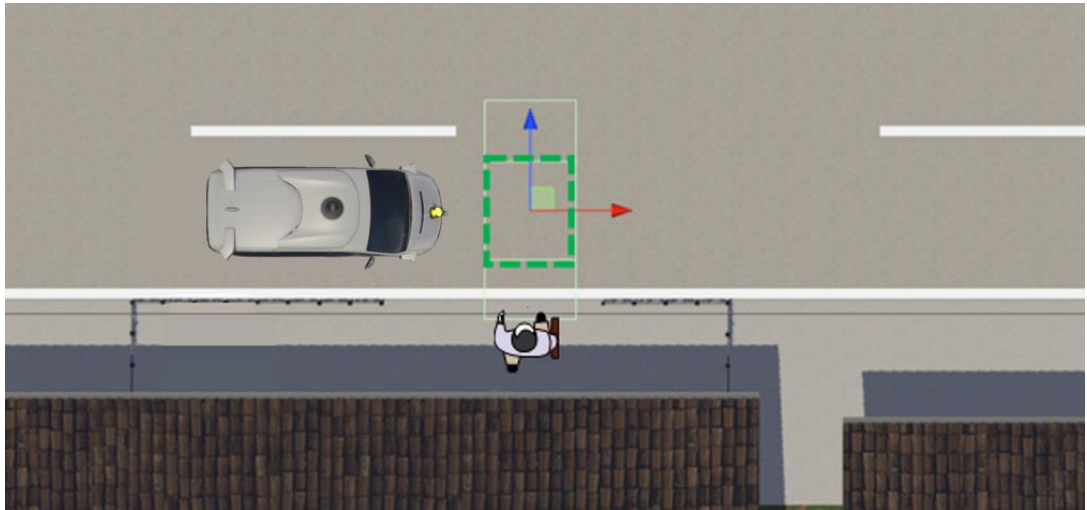

**Figure 3 Conflict Box Design in VR using Unity**

### 3. YB Signal Scenario

This study also introduces a YB signal scenario that incorporates a platoon of non-yielding AVs displaying blue signals in addition to the Y signal scenario. The blue signal concept was developed with the aim of improving traffic operation. Specifically, when a group of vehicles is in a queue, it is beneficial to the traffic system if they can maintain their speed without stopping at unmarked midblock locations. So, the group of vehicles will take priority ongoing and will not stop for the pedestrian. After the platoon of blue vehicles, there will be a few yellow vehicles that will give pedestrians more priority, and behind that would be potentially another platoon of blue vehicles coming (Figure 4).

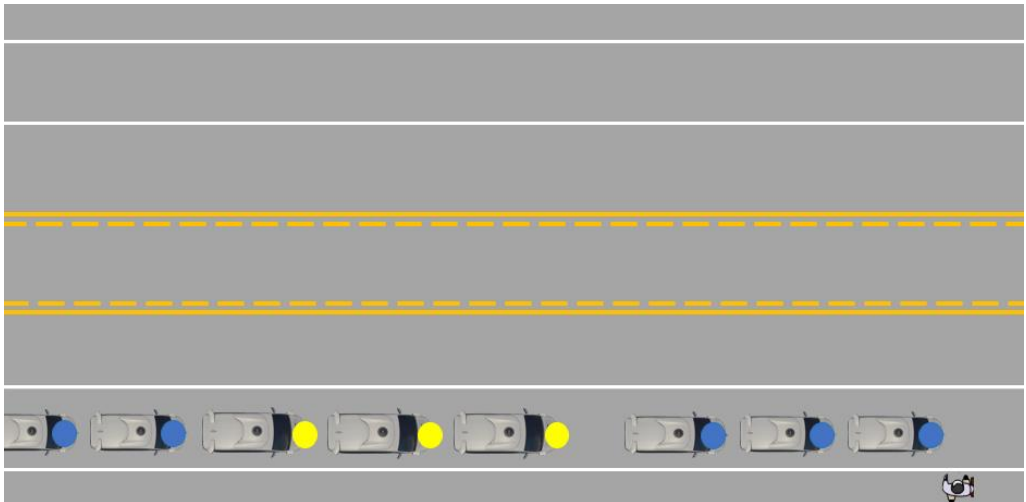

**Figure 4 An Example of YB signal scenario**

*Gap design*

The determination of appropriate gap sizes between AVs is a critical aspect of experiment design. In the present study, it is imperative to identify a gap that is neither too large nor too small and will be deemed acceptable by pedestrians. Prior research examining midblock crossings has reported that no one crossed below 3 seconds, and all crossed over 8 seconds (Schmidt & Färber, 2009). Research also indicated that the near-side gap is considered the critical gap when pedestrians cross at midblock locations (Wang et al., 2010). The literature review also suggested designing multiple experimental conditions with different gaps (Dietrich, Maruhn, et al., 2020). Based on several pilot studies, the researchers determined that a 4.5-second near-side gap and a 3.5-second near-side gap after the middle of the road in Figure 5 would be appropriate. Furthermore, a far-side gap of 15 seconds was implemented to ensure that pedestrians' interactions with the AV in the first lane would not be impacted.

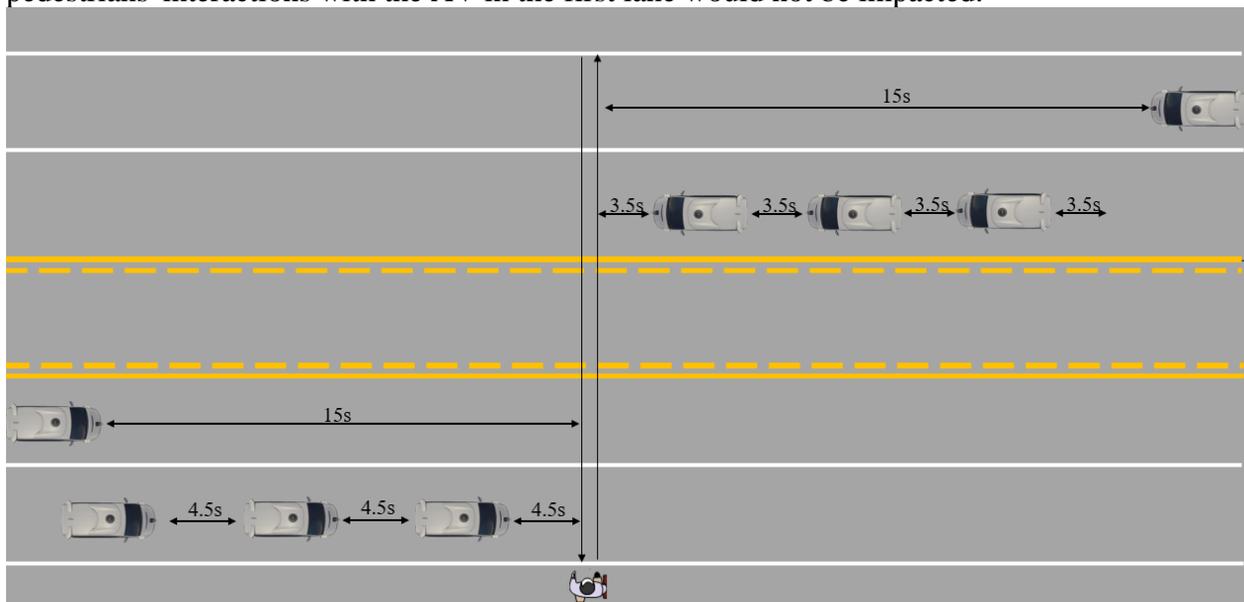

**Figure 5 Gap Design for the AVs in VR**

*Pedestrian Task*

Figure 6 below exhibits one participant in the VR experiment. It is a TWLTL scene with an AV showing a red signal warning the pedestrian. The three groups of pictures show the participant waiting at the curb, waiting in the middle of the TWLTL, and picking up food on the opposite side of the road to complete the task. Pedestrians' task designed in this VR experiment is to cross the unmarked midblock road outside their houses to pick up food from the restaurant across the road, as the nearest crosswalk is too far away. Figure 6 (right) shows that the designed task incorporates an approximately 20 ft distance of walking on the sidewalk before crossing the road to provide a more realistic scenario for the pedestrians. The participants were informed of the presence of AVs on the road. Upon completing the task (picking up food) and returning to the starting point, a new randomized simulation scenario with instructions would automatically generate for the participant, as shown in Figure 7. The pedestrian task allowed for multiple crossings without explicit instructions from researchers, minimizing interference with participants' immersion. Furthermore, participants will focus on their tasks without the realization that their crossing behaviors are of primary interest to the research (Jayaraman et al., 2018).

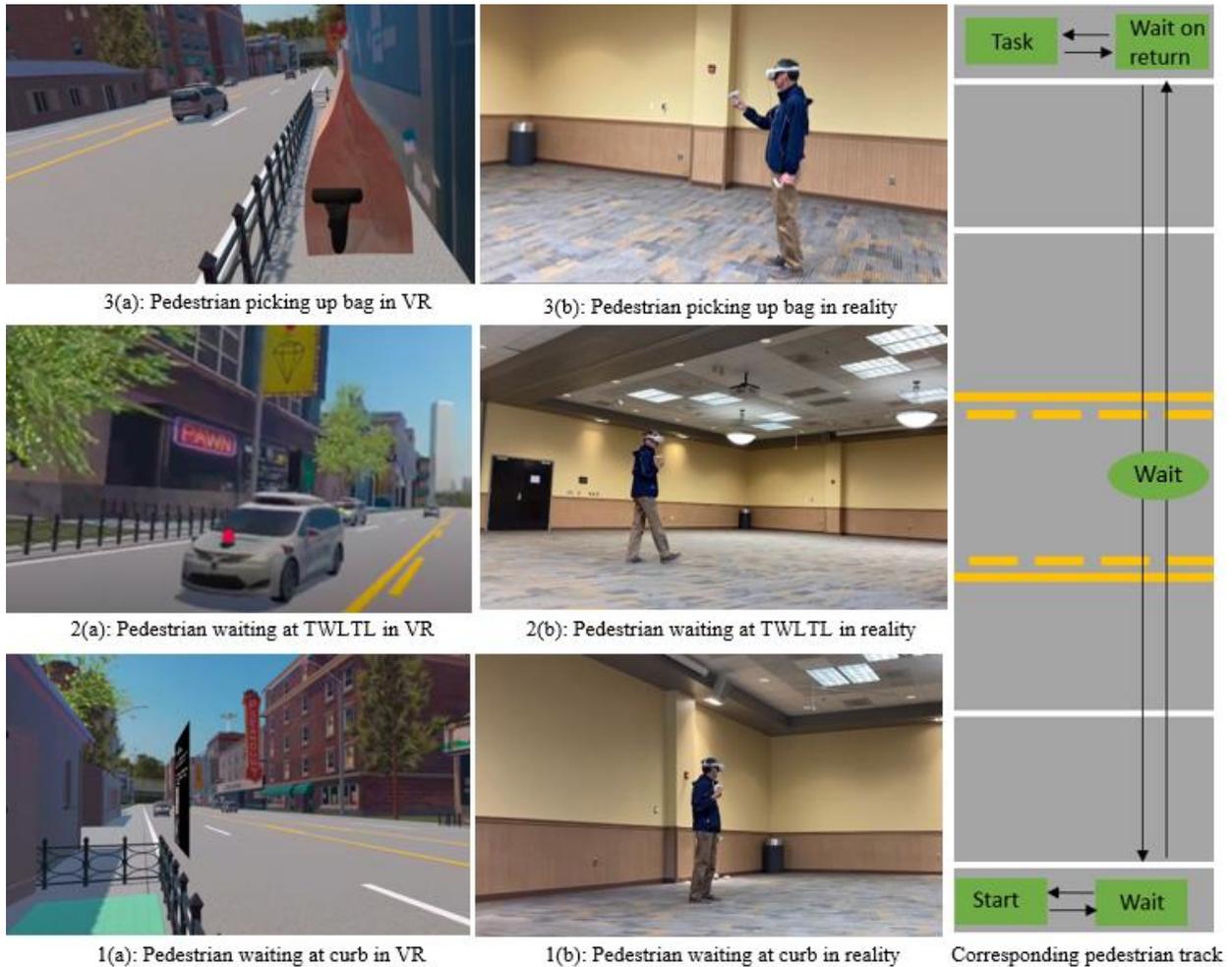

3(a): Pedestrian picking up bag in VR     3(b): Pedestrian picking up bag in reality

2(a): Pedestrian waiting at TWLTL in VR     2(b): Pedestrian waiting at TWLTL in reality

1(a): Pedestrian waiting at curb in VR     1(b): Pedestrian waiting at curb in reality     Corresponding pedestrian track

**Figure 6 Pedestrian in VR (left), Reality (middle), and Pedestrian Walking Track (right)**

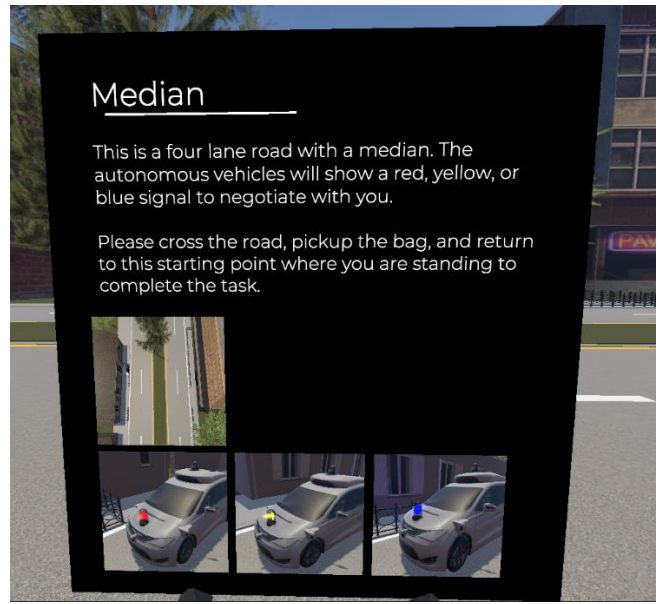

**Figure 7 Instruction Board in VR for Pedestrians**

*Experiment Procedure*

Figure 8 provides an overview of the experiment flow, which involved a two-factor within-subject design consisting of roadway scenes and AV signals. The experiment and accompanying survey required each participant to devote approximately 45 to 90 minutes of their time. Before commencing the experiment, participants provided their consent and completed multiple questionnaire surveys on demographics (age and gender), walking exposure (utilitarian trips, daily walking time, and infrastructure adequacy), and a 20-item pedestrian behavior questionnaire (Deb, Strawderman, et al., 2017) which measures past violations, errors, lapses, aggressive behaviors, and positive behaviors. Participants experienced a training session with the headset. The training was designed to 1) help participants become familiar with completing the task using the VR equipment, thus reducing distraction and promoting immersion; 2) to clarify the meaning of AV signals and behaviors to participants. Research has demonstrated that training enhances participants' understanding of AV signals conveyed through the interface, increasing their confidence in interpreting AV signals and behaviors (Lagström & Victor Malmsten Lundgren, 2015; Lee et al., 2019). During the training, participants received instructions on the meaning of the AV yellow, red, and blue signals and interacted with AVs exhibiting various signals in the first lane. After the training, participants were asked to answer questions to confirm their comprehension of different signals. The research team continued with the experiment only if participants understood the meaning of each AV signal (yellow indicates AV wants to negotiate right-of-way with the pedestrian, red indicates danger and AV lacks sufficient time to stop, and blue indicates AV is in a platoon and will not stop for the pedestrian). Then participants started the experiment with a combination of randomized scenes and AV signals. Each experiment block includes nine randomized scenarios, and each participant finished three repeated blocks (27 scenarios total) to complete the experiments. The research team also monitored participants' simulator sickness using the 16-item simulation sickness questionnaire (SSQ) after training and after each experimental trial (R. S. Kennedy et al., 1993). Every participant received a $20 gift card in compensation for their time.

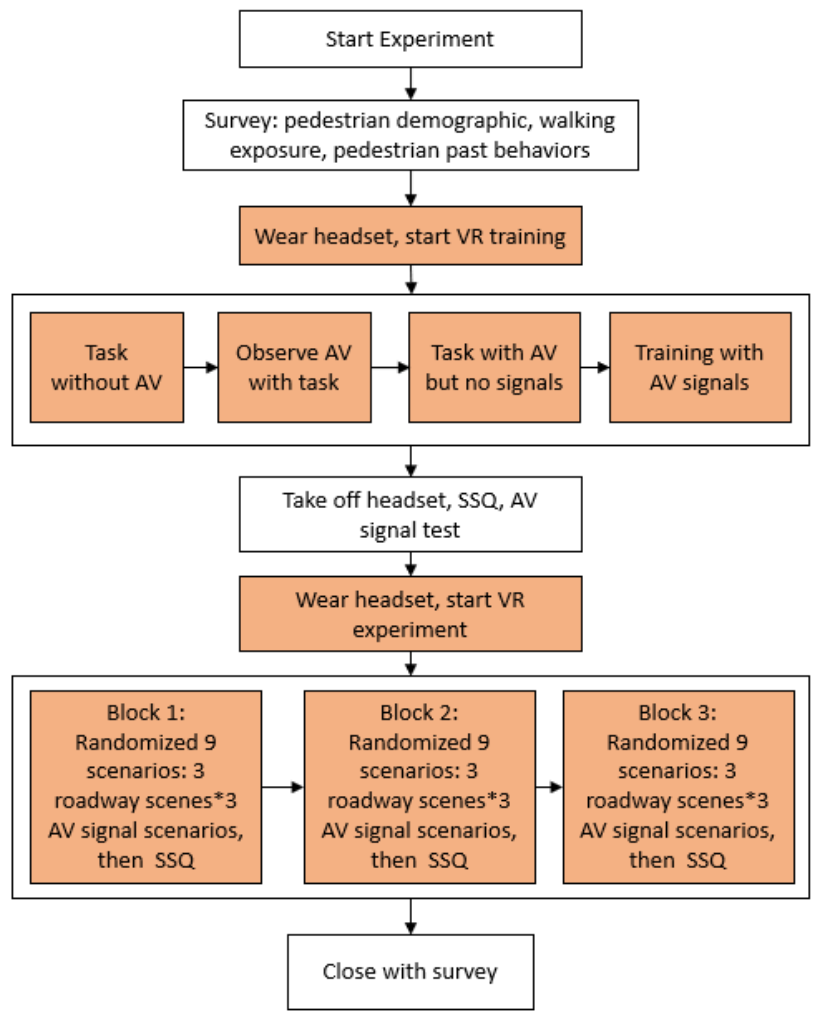

Note: SSQ is simulation sickness questionnaire;
Boxes with color are in VR

**Figure 8 An Overview of the Experimental Procedure**

**Data Collection and Preparation**

The research team collected both VR simulator-based pedestrian behavior data and survey-based responses. A total of 50 participants from Clemson University and the city of Clemson community were recruited, representing a larger sample size compared to previous studies on this topic (Böckle et al., 2017; Camara et al., 2021; Chang et al., 2017; Clercq et al., 2019; Deb et al., 2018; Holländer et al., 2019; Jayaraman et al., 2019; Mahadevan et al., 2019; Pillai, 2017; Stadler et al., 2019). The VR-based pedestrian behavior data captured the position, speed, and head rotation of the pedestrian for each timestamp. Additionally, the simulation recorded the AV positions, speed, and AV signal types for each lane. No participants experienced severe motion sickness or discontinued the experiment. The average overall simulator sickness score of 5.78, with participants spending nearly one hour in the virtual environment, is considered minimal sickness symptoms, as indicated by a previous study (R. , Kennedy et al., 2003).

*Data cleaning*

Each participant completes 27 scenarios, resulting in 27 data files per participant. However, one participant completed only 20 scenarios, while another completed only 18 scenarios as the headsets ran out of battery. Overall, there were 1334 data files collected from the 50 participants. Following data cleaning procedures, three files were excluded from the analysis: two files captured the return trip after the participant completed the task, while the other file recorded the participant staying at the starting point. Ultimately, 1331 data files remained for further analysis.

*Data Preparation*

This study aims to understand pedestrians' diverse crossing behaviors at unmarked midblock locations with respect to different roadway scenes and AV signals and behaviors. Table 1 shows the variables for data analysis. The analysis includes four dependent variables obtained from the VR experiment: waiting time at the curb, waiting time in the middle, total crossing time, and walking time. Taking waiting time at the curb as an example, it starts timing as the pedestrian turns from the direction of walking on the sidewalk to the direction of crossing the road; it ends timing as the pedestrian leaves the curb. It's worth noting that pedestrian hesitation is also considered when calculating the waiting time. For example, pedestrian hesitation happens when the pedestrian leaves the curb, enters the first lane, but returns to the curb again. This hesitation period is included as the time the pedestrian spent waiting at the curb. A similar procedure is applied to analyze waiting time in the middle.

Based on the experiment design, the main independent variables include scenes (undivided, TWLTL, and median) and AV signals (No signal, Y signal, and YB signal), as shown in Table 1. Other independent variables, including pedestrian demographics, walking exposure, and pedestrian past behavior, were also explored in the analysis.

**Table 1 Variables for Modeling**

| Dependent Variables | | | |
|---|---|---|---|
| **Variable name** | Definition | | |
| Waiting time at the curb | Time the pedestrian spends at the curb waiting to cross the road | | |
| Waiting time in the middle | Time pedestrian spends in the middle of the road waiting to cross to the other side of the road | | |
| Total crossing time | Total time pedestrian spends crossing the road from the nearest curb to the farthest curb | | |
| Walking time | Time pedestrian spends walking on the road, which equals to total crossing time deducting the waiting time at the curb and waiting time in the middle | | |
| **Independent Variables** | | | |
| **Variable name** | Levels | Annotation | Explanation |
| **Scene** | 3 | Undivided | Undivided four-lane road |
| | | TWLTL | Four-lane road with a TWLTL |
| | | Median | Four-lane road with a median |
| **Signal** | 3 | No Signal | AV with negotiation behavior showing no signal |
| | | Y Signal | AV with negotiation behavior showing a yellow signal |
| | | YB Signal | AV with non-stopping behavior showing a blue signal added to the Y signal scenario |
| **Demographics** | 5 | Age | Five groups: 18-29; 30-39; 40-49; 50-59; 60+ |
| | 2 | Gender | Male, female |

| **Walking exposure** | 5 | Walking time | The range for total daily walking time (min) |
| | 4 | Walking trips | From never to frequently, four levels |
| | 5 | Proper infrastructure | From never to always, five levels |
| **Pedestrian past behavior** | NA | Violations | Four questions averaged as past violations; |
| | | Errors | Four questions averaged as past errors; |
| | | Lapses | Four questions averaged as past lapses; |
| | | Aggressive | Four questions averaged as past aggressive behavior |
| | | Positive | Four questions averaged as past positive behavior |

## Model

A linear mixed model is used to allow for correlation between the response variables within the same participant (Jin et al., 2022; Vitorino et al., 2020). Each participant completes multiple crossing scenarios in the VR environment, so the response variables are measured repeatedly within the same participant. Therefore, response variables are correlated within a participant. The linear mixed model is represented as follows,

$$\boldsymbol{y} = \boldsymbol{X\beta} + \boldsymbol{Zu} + \boldsymbol{\varepsilon}$$

Where, $\boldsymbol{y}$ is a $N \times 1$ vector of the dependent variable. $\boldsymbol{X}$ represents a $N \times p$ design matrix of $p$ explanatory variables. $N$ is the number of observations in the dataset used in the model and defined as

$$N = \sum_{i=1}^{q} n_i;$$

$n_i$ is the number of observations within the participant $i$; $q$ is the number of participants in the dataset; $\boldsymbol{\beta}$ represents a $p \times 1$ vector of the fixed effects regression coefficients; $\boldsymbol{Z}$ is the $N \times q$ matrix of $q$ random effects; $u$ is a $q \times 1$ vector of the random effects. $\boldsymbol{u} \sim N(0, \boldsymbol{G})$; $\boldsymbol{G}$ is the variance-covariance matrix of the random effects. $\boldsymbol{\varepsilon}$ is a $N \times 1$ vector of the residuals.

ANOVA is used to test if an interaction effect between two main factors exists. When the interaction effect is significant, least-squares means can be estimated to perform multiple comparisons. When interactions are not significant, main effects are investigated.

R software is utilized to run linear mixed model using "lmr" package and ANOVA test is also conducted in R (R Core Team, 2013).

## RESULTS

The results part first presented the descriptive statistics of the four dependent variables used in the study: waiting time at the curb, waiting time in the middle, walking time, and total crossing time. Then the effects of the main factors, Scene and AV signal, were explored using a linear mixed model and ANOVA test. Pairwise differences in Scenes and AV signals were also presented for detailed comparisons, followed by the interaction effects between Scene and Signal for the waiting time in the middle. Effects for pedestrian demographics, walking exposures, and pedestrian past behavior collected from the survey were presented at the end of this section.

## Descriptive Statistics

Data collected from the VR experiment showed that, on average, participants spent 6.06 seconds waiting at the curb and 4.26 seconds in the middle of the multilane. Participants spent 11.50 seconds on average walking four lanes. And the average total crossing time, including the time spent walking and waiting, was 21.82 seconds. Figure 9 (top left) shows that compared to the median and TWLTL scenes, participants spent more time waiting at the curb and much less time in the middle of the road at the undivided scene. Further, participants spent more time waiting in the middle for the median scene and spent more time waiting at the curb for the TWLTL scene. Figure 9 (top right) shows that the Y signal had the shortest waiting time at both the curb and middle. Figure 9 (bottom) shows the walking time by scene or signal did not differ much. And participants spent the least time in total crossing the road in Y signal scenarios (Figure 9 bottom right). More statistical test results are presented in the next section.

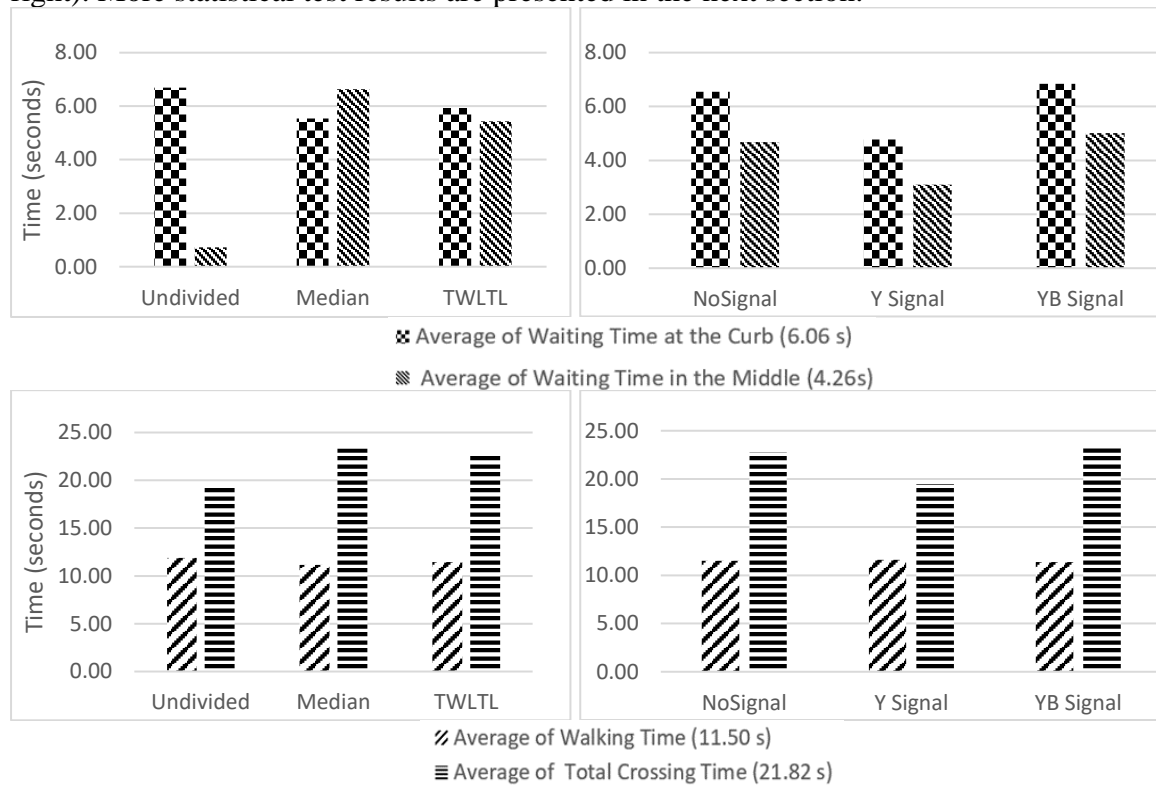

**Figure 9 Descriptive Statistics of the Dependent Variables (top– waiting time at curb and middle of the road, bottom –walking time, and total crossing time) by Scene or Signal**

**Effects of Main Factors**

Table 2 first presents the significance of the main factors: Scene and AV signal. The findings indicate that Scene significantly affects all four dependent variables (note 1), and Signal significantly affects all but the walking time (note 2). Table 2 further provides the pairwise differences of scenes, followed by the pairwise differences of different signals. Additionally, the analysis reveals that the interaction between Scene and Signal is only significant for waiting time in the middle of the road (note 3). The next section will discuss the interaction effects between Scene and Signal.

**Table 2 Effects of Main Factors**

| Main factors Significance (ANOVA) | | | | | | | | |
|---|---|---|---|---|---|---|---|---|
| | Waiting time at the curb | | Waiting time in the middle | | Walking time | | Total crossing time | |
| | F value | Pr(>F) | F value | Pr(>F) | F value | Pr(>F) | F value | Pr(>F) |
| [1]Scene | 6.041 | **0.002** | 222.174 | **<.0001** | 6.696 | **0.001** | 38.336 | **<.0001** |
| [2]Signal | 22.179 | **<0.001** | 23.624 | **<.0001** | 0.505 | 0.604 | 33.6770 | **<.0001** |

| [3]Scene*Signal | 0.309 | 0.872 | 4.563 | **0.0012** | 0.403 | 0.807 | 1.8753 | 0.1124 |
|---|---|---|---|---|---|---|---|---|

| **Pairwise Differences of Scenes** | | | | | | | | |
|---|---|---|---|---|---|---|---|---|
| | Waiting time at the curb | | Waiting time in the middle | | Walking time | | Total crossing time | |
| | mean | p-value | mean | p-value | mean | p-value | mean | p-value |
| [4]Undivided - Median | 1.154 | **0.0006** | -5.92 | **<.0001** | 0.735 | **0.0003** | -4.034 | **<.0001** |
| [5]Undivided - TWLTL | 0.759 | **0.0242** | -4.73 | **<.0001** | 0.454 | **0.025** | -3.512 | **<.0001** |
| [6]Median - TWLTL | -0.395 | 0.2410 | 1.19 | **0.0001** | -0.281 | 0.1660 | 0.522 | 0.2981 |

| **Pairwise Differences of Signal** | | | | | | | | |
|---|---|---|---|---|---|---|---|---|
| | Waiting time at the curb | | Waiting time in the middle | | Walking time | | Total crossing time | |
| | mean | p-value | mean | p-value | mean | p-value | mean | p-value |
| [7]No signal - Y Signal | 1.771 | **<.0001** | 1.58 | **<.0001** | -0.072 | 0.7243 | 3.28 | **<.0001** |
| [8]No signal - YB Signal | -0.308 | 0.3599 | -0.33 | 0.2665 | 0.129 | 0.5241 | -0.51 | 0.3092 |
| [9]Y Signal - YB Signal | -2.080 | **<.0001** | -1.91 | **<.0001** | 0.201 | 0.3217 | -3.79 | **<.0001** |

Note: Bold results are significant at a 0.05 level

Note: Numbers 1-9 marked are the numbers that are called out for results explanation

For the pairwise differences in scenes, taking note 4 as an example, the positive mean difference value of 1.154 indicates that participants waited 1.154 seconds longer at the curb in the undivided scene than in the median scene. Notes 4-5 show that participants in the undivided scene spent a longer time waiting at the curb and walking on the road than in the median and TWLTL scenes. However, they spent a shorter time waiting in the middle of the road. The longer walking time on the road implies that participants spent more time interacting with AVs while walking in the undivided scene compared to the median and TWLTL scenes. Note 6 shows that only the waiting time in the middle of the road is longer in the median scene compared to the TWLTL scene at a 0.05 significance level. The finding of note 6 implies that pedestrians exhibited different waiting behaviors between TWLTL and the median while waiting in the middle of the road, but not at the curb or while walking on the road.

Table 2 shows Signal significantly affects all dependent variables except for the walking time (note 2). The pairwise difference of signals also shows that there are no significant differences between any of the AV signals for walking time, implying that the AV signal has no effect on pedestrian walking behavior after crossing begins. It's also found that except for the walking time, there are always significant differences in waiting time at the curb, waiting time in the middle, and the total crossing time between the No signal and Y signal (note 7). Note 7 shows that compared to the No signal scenario, adding the Y signal indication for AVs resulted in a reduction of pedestrian waiting time at the curb by 1.771 seconds and waiting time in the middle by 1.58 seconds at a significance level of 0.05. Thus, the addition of the Y signal encouraged pedestrians to make crossing decisions faster compared to negotiation without a signal indication. Note 8 indicates that No signal and YB signal are not significantly different. Note 9 shows that compared to the Y signal, in the YB signal scenario, participants waited 2.08 seconds longer at the curb, 1.91 seconds longer at the middle, and 3.79 seconds longer for the total crossing time, with all differences at a 0.05 significance level. The differences imply that when some AVs in the traffic show non-stopping behavior, pedestrians yield both at the curb and in the middle of the road.

**Interaction Effects Between Scene and Signal**

Table 2 (note 3) shows significant interaction effects between Scene and Signal for waiting time in the middle of the road. This section discusses the interaction effects. Figure 10 compares the effects of the combination of three scenes and three AV signals for waiting time in the middle of the road.

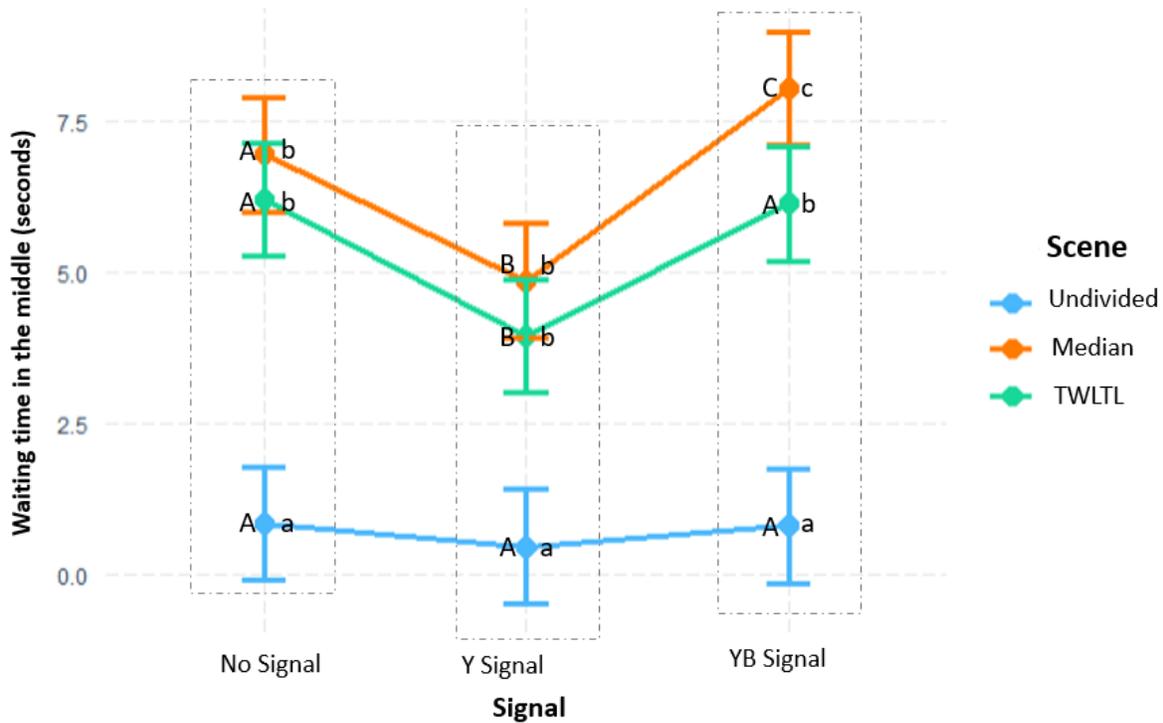

Note: Means follow by a common capital letter are not statistically significant within a scene at a 0.05 level
Means follow by a common lowercase letter are not statistically significant within a signal at a 0.05 level

**Figure 10 Results of Pairwise Differences of Scene and Signal**

In Figure 10, the same color represents results in the same scene, and each grey dotted box represents results of the same AV signal. In Figure 10, means followed by the capital letter are not statistically significant within a scene at a 0.05 level. Means followed by a common lowercase letter are not statistically significant within an AV signal. Taking the green line (TWLTL scene) as an example, capital letters A and B indicate statistically significant differences between No signal and Y signal, Y signal and YB signal. But capital letters A and A of the green line imply that No signal and YB signals are not statistically significantly different. The blue line in Figure 10 shows that for the undivided scene, the differences between the three signals are not statistically significant. The blue line results imply that participants' behavior in the middle of the undivided road does not change between all AV signals. The green line in Figure 10 shows that for the TWLTL scene, there are significant differences between the No signal and Y signal, Y and YB signal, but not between No signal and YB signal. However, the orange line in Figure 10 shows significant differences between all three signal types. For both TWLTL and median, pedestrians spent less time waiting in the middle when the AV gave a Y signal than No signal; the YB signal increased pedestrians waiting time in the middle compared to the Y signal. Although note 8 in Table 2 shows that No signal and YB signal are not significantly different without considering the interaction effect between Scene and Signal, Figure 10 shows that compared to No signal, the waiting time in the middle increased for the YB signal in the median scene (not TWLTL and undivided scene). The left and middle grey dotted boxes in Figure 10 reveal that No signal and Y signal have significantly different effects between undivided and median, undivided and TWLTL, but No significant differences between median and TWLTL. But for the group of YB signals, the right box in Figure 10 shows that pedestrians waited longer at the median than at the TWLTL at a statistical significance level. The results also indicate that pedestrians feel more comfortable staying longer at the median and yield more to AVs in the YB signal scenario, with some AVs showing non-stopping behavior.

**Effects of Demographics**

Table 3 below displays the effects of the demographic factors age [age 18-29: 33 people, 30-39: 6 people, 40-49: 6 people, 50-59: 1 people, 60+: 4 people] and gender [male: 31 and female: 19] collected from the survey before the VR experiment. Age groups 40-49 and 50-59 were combined as one group, 40-59, because there was only one participant in the group of 50-59. The tests were done across all signals and scene scenarios and gave the overall effects of age and gender on the four dependent variables.

Table 3 shows that age is significant for waiting time at the curb, waiting time in the middle, and the total crossing time, but not for the walking time (note 1 in Table 3). Pairwise differences in age groups show that no age groups differ significantly in walking time. But some age groups significantly differ in the waiting time at the curb, waiting time in the middle, and total crossing time. For example, note 3 shows that compared to age group 18-29, group 40-59 waited 3.693 seconds longer at the curb, 1.928 seconds longer in the middle, and spent overall 4.193 seconds more crossing the multilane, with all results at a 0.05 significance level. Age groups 60+ waited longer in the middle of the road and overall spent more time crossing the road compared to age groups 18-29 (note 4) and 30-39 (note 5). Gender is only significant for waiting time in the middle and the total crossing time (note 2). On average, females wait 1.66 seconds longer than males in the middle of the road, but their walking time is not significantly different from men (note 6).

**Table 3 Effects of Demographics**

| Demographic factors Significance (ANOVA) | | | | | | | | |
|---|---|---|---|---|---|---|---|---|
| | Waiting time at the curb | | Waiting time in the middle | | Walking time | | Total crossing time | |
| | F value | Pr (>F) | F value | Pr (>F) | F value | Pr (>F) | F value | Pr (>F) |
| [1]Age | 3.582 | **0.0207** | 5.9368 | **0.0016** | 0.8664 | 0.4653 | 3.610 | **0.0201** |
| [2]Gender | 3.191 | 0.0803 | 6.3691 | **0.0149** | 0.0439 | 0.8349 | 5.693 | **0.0210** |
| **Pairwise Differences of Age** | | | | | | | | |
| | Waiting time at the curb | | Waiting time in the middle | | Walking time | | Total crossing time | |
| | mean | p-value | mean | p-value | mean | p-value | mean | p-value |
| (18-29) - (30-39) | -1.264 | 0.3567 | -0.323 | 0.7282 | 0.6995 | 0.4864 | -0.888 | 0.6821 |
| [3](18-29) - (40-59) | -3.693 | **0.0057** | -1.928 | **0.0307** | 1.4298 | 0.1328 | -4.193 | **0.0434** |
| [4](18-29) - (60+) | -3.04 | 0.0668 | -4.219 | **0.0004** | 0.0505 | 0.9663 | -7.209 | **0.0073** |
| (30-39) - (40-59) | -2.43 | 0.1599 | -1.605 | 0.1716 | 0.7303 | 0.5617 | -3.305 | 0.227 |
| [5](30-39) - (60+) | -1.776 | 0.3728 | -3.896 | **0.0057** | -0.649 | 0.6565 | -6.321 | **0.0494** |
| (40-59) - (60+) | 0.653 | 0.7347 | -2.291 | 0.0853 | -1.3793 | 0.3322 | -3.016 | 0.3265 |
| **Pairwise Differences of Gender** | | | | | | | | |
| | Waiting time at the curb | | Waiting time in the middle | | Walking time | | Total crossing time | |
| | mean | p-value | mean | p-value | mean | p-value | mean | p-value |
| [6]Female - Male | 1.68 | 0.0804 | 1.66 | **0.0150** | -0.072 | 0.7243 | 3.47 | **0.021** |

Note: Bold results are significant at a 0.05 level
Note: Numbers 1-6 marked at the beginning of the rows are the numbers that are called out as notes for results explanation

**Effects of Past Behaviors and Walking Exposure**

Other variables collected from the survey were also tested in the models. Past crossing behaviors (e.g., violations, errors, lapses, and aggressive and positive behavior) were not statistically significant in the model. Past walking exposures, such as utilitarian walking trips, daily walking time, and walking infrastructure adequacy, were also insignificant in the model. This suggests that pedestrians' behavior interacting with AVs is primarily influenced by their experience in the VR experiment, as the interaction with AVs is still a new and unfamiliar experience for pedestrians. Specifically, the results suggest that pedestrians' past walking exposure or behavior interacting with human-driven vehicles do not significantly affect their behavior interacting with AVs. Thus, pedestrians' behavior interacting with AVs is more dependent on their experience with AVs than their past experience.

**CONCLUSION AND DISCUSSION**

One of the main challenges AVs will face is interacting with pedestrians, especially at unmarked midblock locations on multilane roads where the right-of-way is unspecified. Understanding pedestrians' behavioral responses to AVs and infrastructure design, as well as the development of effective AV operational and communication strategies, can help AVs respond accordingly. This study used VR to simulate an urban/suburban midblock environment where pedestrians interact with AVs as they cross a four-lane arterial roadway.

Three roadway scenes were designed in VR: a 4-lane undivided, a 4-lane with a TWLTL, and a 4-lane with a median. Three AV signal scenarios were included in the experiment: a No signal scenario, a Y signal scenario, and a YB signal scenario. In the No signal scenario, AVs negotiate the right-of-way with pedestrians without any signal indication. In the Y signal scenario, AVs negotiate the right-of-way with pedestrians through a yellow signal indication portrayed to pedestrians. In the YB signal scenario, a platoon of AVs displaying blue signals indicating that they will not stop for pedestrians is added to the Y signal scenario. Outcomes of interest included whether pedestrian behavior changes with the provision of a centerline refuge area or when pedestrians experience different AV operations portrayed to pedestrians through signaling communications. Additionally, researchers were also interested in determining any interaction effects between the roadway infrastructure and the AV operations.

**Effects of Roadway Scene**
Results show that roadway scenes significantly impact pedestrians' unmarked midblock crossing behavior, including the waiting time at the curb, waiting time in the middle, walking time, and total crossing time. On average, when pedestrians cross multilane roads at unmarked midblock locations while interacting with AV, they spend more time waiting at the curb than in the middle of the road. This difference is particularly pronounced in the 4-lane undivided scene, where no location of refuge or pause exists in the middle of the road. While researchers expected this scenario to have the fastest walking time, it had the slowest. Authors presume this is due to decision-making while walking slowing the process down overall. Moreover, results also show that curb waiting time for TWLTL and median had the same response, but the two scenes had significantly different waiting times in the middle of the road, with the median being longer overall. These results indicate that infrastructure design will continue to impact pedestrian behavior and should be considered in future AV system design.

**Effects of AV Behavior and Signal**
AV behavior and signals significantly impact pedestrians' unmarked midblock crossing behavior, including the waiting time at the curb, waiting time in the middle, and the total crossing time, but not the walking time. The most notable effect was observed with the Y-signal indication, which conveyed that the AV had detected the pedestrian and would stop if the pedestrian entered the conflict box on yellow. Pedestrians immediately responded to the Y signal by entering the road and assuming right-of-way, demonstrating their intent to take advantage of the risk-averse AV detection and response system. Compared with the No signal operation, the Y-signal significantly reduced pedestrian waiting time and total crossing time. The addition of a blue signal to the yellow scenario (YB scenario) enabled more aggressive AV operations with platoons of vehicles, indicating that they would not stop for pedestrians to prioritize vehicle mobility. The YB signal scenario significantly increased the pedestrian waiting time at the curb and middle. So, when some AVs in the traffic show more aggressive behavior, pedestrians tend to yield both at the curb and in the middle of the road. These results suggest that a more controlled crossing can be achieved with the addition of a platoon of non-yielding AVs. AV operational design, training, and enforcement will all be required for efficient and effective transportation systems.

**Interaction Effects Between Scene and Signal**
Results also show that the interaction of signal and scene is only significant for waiting time in the middle of the road. Researchers found that participants' behavior in the middle of the 4-lane undivided road remains consistent across all AV signals tested. In terms of middle waiting time, the introduction of non-stopping platoons with blue signals had the most substantial impact on the median roadway type, whereas the impact was least pronounced on the 4-lane undivided road. This discrepancy may be attributed to pedestrians on a multilane road with a median feeling more secure and confident in waiting, observing, and yielding to an AV displaying more aggressive behavior. Conversely, the absence of a refuge spot may lead to a continuation of the crossing movement.

**Effects of other factors**
This study also investigates the impact of demographics on pedestrian behavior. The results indicate that older individuals tend to wait longer before making crossing decisions, particularly in the middle of the road. Females tend to wait longer than males in the middle of the road for decision-making, but no significant difference is observed in their walking time. To further advance the research, it would be intriguing to examine the compliance of different age and gender groups with the designed signals, such as identifying the age or gender group that violated the blue signals the most. In addition, other variables, such as past walking exposures and pedestrian past behaviors, were also tested in the models. The results reveal that pedestrians' past walking exposure or behavior interacting with human-driven vehicles does not significantly impact their behavior when interacting with AVs. Thus, pedestrian behavior when interacting with AVs is more reliant on their experience with AVs than their past encounters.

To conclude, this paper studies pedestrian behavior interacting with AVs at unmarked midblock locations on a four-lane road. Different types of multilane roads and AV behavior and signals are tested in VR to explore the association between pedestrian behavior and roadway infrastructure and AV behavior and communication strategies. Notably, pedestrians experienced multiple AV interactions in the same scenario, with AVs being programmed to negotiate right-of-way with pedestrians to facilitate real-time, back-and-forth interaction. Additionally, this paper explores the interaction between the roadway infrastructure and AV behavior and signals. As we are designing new AV algorithms, we can test how pedestrians react to different AV behaviors. We also conclude that infrastructure design will continue to impact pedestrian behavior and should be considered in future AV system design.

## ACKNOWLEDGMENTS

We would thank Dr. Mashrur "Ronnie" Chowdhury at Clemson University for the funding of data collection.

## AUTHOR CONTRIBUTIONS

The authors confirm their contribution to the paper as follows: study conception: F. Zou, J. Ogle, W. Jin; VR experiment design and data collection: F. Zou, J. Ogle, D. Petty, A. Robb, W. Jin, P. Gerard; analysis and interpretation of results: F. Zou, J. Ogle, W. Jin, P. Gerard, A. Robb; draft manuscript preparation: F. Zou, J. Ogle, W. Jin, P. Gerard. All authors reviewed the results and approved the final version of the manuscript.